%% file: main.tex
\documentclass{article}

\usepackage{microtype}
\usepackage{graphicx}
\usepackage{subcaption}
\usepackage{booktabs} 

\usepackage[accepted]{icml2020}
\input{settings.tex}
\icmltitlerunning{Discovering Invariances in Healthcare Neural Networks}

\begin{document}
\twocolumn[
\icmltitle{Discovering Invariances in Healthcare Neural Networks}
\begin{icmlauthorlist}
\icmlauthor{Mohammad Taha Bahadori}{}
\icmlauthor{Layne C. Price}{}
\end{icmlauthorlist}
\icmlcorrespondingauthor{M. T. Bahadori}{bahadorm@amazon.com}
\vskip 0.3in
]

\begin{abstract}
    \input{abst.tex}
\end{abstract}

\section{Introduction}
\input{intro.tex}

\section{Invariance Discovery Methodology}
\input{method.tex}
\input{analysis.tex}

\section{Experiments}
\input{exp.tex}

\section{Conclusion and Future Works}
\input{conc.tex}

\section*{Acknowledgements}
The authors are thankful of Edward Choi for helpful discussions and suggestions. 
The authors also thank Daniel Navarro, RN, for discussions about our results from a clinical prospective.

\bibliography{references}
\bibliographystyle{icml2020}
\balance
\newpage
\clearpage

\appendix
\section{Temporal Basis Functions}
\label{sec:basis}
\input{basis.tex}

\section{Proof of the Theorems}
\label{sec:proofs}
\input{proofs.tex}

\onecolumn
\section{Details of the Models}
\label{sec:models}
\input{details.tex}
\label{sec:names}
\input{names.tex}
\end{document}

%% file: settings.tex
\usepackage[utf8]{inputenc} 
\usepackage[T1]{fontenc}    
\usepackage{microtype}
\usepackage{graphicx}
\usepackage{longtable}
\usepackage{subcaption}
\usepackage{booktabs} 
\usepackage{amsfonts}
\usepackage{amsthm}
\usepackage{nicefrac}
\usepackage{enumitem}
\usepackage{balance}
\setlist{leftmargin=1.5mm}

\usepackage{bm,amsmath,amssymb,amsfonts}
\usepackage{longtable,lscape}
\usepackage{booktabs}
\RequirePackage{color}
\newtheorem{theorem}{Theorem}

\newtheorem{definition}{Definition}
\definecolor{mydarkblue}{rgb}{0,0.08,0.45}
\usepackage{array}

\usepackage{url}
\usepackage{multirow}
\DeclareCaptionType{Algorithm}

\DeclareMathOperator*{\argmin}{argmin}

\usepackage[utf8]{inputenc} 
\usepackage[T1]{fontenc}    
\usepackage{booktabs}       
\usepackage{amsfonts}       
\usepackage{nicefrac}       

\usepackage{algorithm}
\usepackage{algorithmic}
\usepackage{bm}
\usepackage{enumitem}

%% file: abst.tex
We study the invariance characteristics of pre-trained predictive models by empirically learning transformations on the input that leave the prediction function approximately unchanged. To learn invariant transformations, we minimize the Wasserstein distance between the predictive distribution conditioned on the data instances and the predictive distribution conditioned on the transformed data instances. To avoid finding degenerate or perturbative transformations, we add a similarity regularization to discourage similarity between the data and its transformed values. We theoretically analyze the correctness of the algorithm and the structure of the solutions. Applying the proposed technique to clinical time series data, we discover variables that commonly-used LSTM models do not rely on for their prediction, especially when the LSTM is trained to be adversarially robust. We also analyze the invariances of BioBERT on clinical notes and discover words that it is invariant to.

%% file: intro.tex
Understanding the invariant properties of a complex neural network  can help analyze its robustness and explain its predictions. Consequently, it can also help us debug the network and detect unwanted behavior \citep{jacobsen2019excessive,gopinath2019finding}. The need for model inspection tools is strongly felt especially in sensitive applications such as healthcare. Moreover, analyzing invariances of healthcare models can uncover insights about the model's function that is unique to health data and guide the model development process. 
In this work, we provide a computationally efficient framework for discovering invariances in predictive models that are differentiable with respect to their inputs. 

Our objective is to  discover transformation, feature, and patterns invariances of a target model. We solve these problems by solving the transformation invariance identification problem. We learn a transformation that minimizes the distance between a pre-trained model's conditional distribution on the data instances and the conditional distribution on the transformed data instances, while simultaneously favoring transformations that are dissimilar from the original data. We choose the Wasserstein distance \cite{villani2008optimal} to  simplify the loss function and add a similarity-based regularization term to encourage non-degenerate and non-perturbative solutions. The loss function is differentiable with respect to the parameters of the transformation; thus we can learn it using auto-differentiation and stochastic gradient descent algorithms. We call our framework the Model INvariance Discovery (MIND) framework.

The MIND framework provides a \textit{global} (dataset-wide) explanation of the model behavior. 
To quantify the amount of invariance, we measure how different, in terms of correlation coefficient, we can make each feature without significantly changing the predictive distribution. 
We further show that by choosing an affine transformation in our framework, we can study the impact of varying each feature in the data and assign each feature a normalized ``\textit{MIND Score}'' between 0 and 1, which assesses the model's sensitivity to that feature. The affine framework allows us to find the invariance of the models to the patterns in the data. 

To analyze MIND's theoretical correctness, we define ``\textit{Weak Invariance}'' for any arbitrary model that is differentiable with respect to its inputs. We show that with an appropriate choice of the similarity regularization parameter, MIND scores correctly discover all of the features that the model is weakly invariant to. We further analyze the structure of the solution for a simpler but related problem and quantify the impact of co-linearity in accuracy of our algorithm.

In comparison to attribution techniques (cf. \cite{Murdoch22071} and the references therein) that are primarily local to existing elements in the data, the MIND framework provides a global summary of model invariances with respect to a reference dataset, with theoretical guarantees.  
We demonstrate the MIND framework summarization capabilities on two concrete examples: we analyze the efficacy of indicator variables in handling missing values in clinical time series; and we characterize the impact of adversarial training in the invariance behavior of  LSTMs.

We perform extensive experiments on two types of healthcare data and two valuable prediction tasks defined on the publicly available MIMIC-III dataset \citep{johnson2016mimic}. First, we use the benchmark in-hospital mortality prediction task from multivariate clinical time series \citep{harutyunyan2017multitask} and compare the invariances learned by  popular LSTM and Transformer models.  In the second set of experiments, we fine-tune the publicly available BioBERT \citep{lee2019biobert} on clinical notes to predict length of stay and analyze its invariant properties. Finally, we show that MIND performs comparatively well in the model randomization sanity checks proposed by \citet{adebayo2018sanity}.

%% file: method.tex
Suppose we have a dataset $\mathcal{D} = \{(X_i, y_i)\}$ for $i=1, \ldots, n$ of feature-label pairs $(X, y) \in \mathbb{X}\times \mathbb{Y}$ and a pre-trained predictive model $p_\theta(y | X): \mathbb{X}\times \mathbb{Y} \mapsto [0,1]$, where the real parameters $\theta$ are fixed. For example, the features $X$ might represent a patient's time series of vital signs; the label $y$ can be their status (deceased, alive) at the time of observation; and $p_\theta$ is the predicted probability of the patient's status given their vital signs.
Our objective is to find a transformation $T_{\bm{\phi}}: \mathbb{X} \mapsto \mathbb{X}$ such that the prediction of $y$ does not change if it is conditioned on $X$ or $T_{\bm{\phi}}(X)$. 
In practice, we are interested in finding a set of transformations parameterized by real values $\bm{\phi}$.
Formally, we find a transformation $T_{\bm{\phi}}$ by optimizing the parameters $\bm{\phi}$ such that the following equation is satisfied:
\begin{equation*}
    p_\theta(y|X) = p_\theta(y|T_{\bm{\phi}}(X)).
\end{equation*}
To enforce the equality in the above equation, we minimize a distance function $D: [0, 1]\times[0, 1] \mapsto \mathbb{R}_{\geq 0}$ between $p_\theta(y|X) $ and $ p_\theta(y|T_{\bm{\phi}}(X))$. To prevent the trivial solution of $T_{\bm{\phi}}$ collapsing to the identity transformation, we add a regularization term as follows:
\begin{equation}
    \widehat{\bm{\phi}} = \argmin_{\bm{\phi}} D\left[p_{\bm{\theta}}(y| X), p_{\bm{\theta}}(y| 
    T_\phi(X))\right] + \lambda S(X, T_\phi(X))
\label{eq:subbed}
\end{equation}
where $S: \mathbb{X}\times \mathbb{X} \mapsto \mathbb{R}$ is a similarity regularization function and $\lambda > 0$ is the penalization coefficient.

\subsection{Computationally Efficient Solutions}
We make the following choices to simplify the computation of Eq.~\eqref{eq:subbed}:

\paragraph{The distance metric $\bm{D}$:} We use the Wasserstein-1 distance as a robust metric for measuring the distance between two distributions \citep{villani2008optimal,cuturi2014fast}. While there are efficient ways for computing the Wasserstein-1 distance \citep{cuturi2013sinkhorn,frogner2015learning}, we can use simple estimation schemes in this paper because our target prediction tasks have a low-dimensional label-space that follows a few simple distributions.  
Namely, the Wasserstein-1 distance reduces to
\begin{equation*}
    W_1(p_\theta, p_{\theta'}) = | f_\theta - f_{\theta '} |
\end{equation*}
in the following circumstances: (1) binary classification tasks when we model the output with a Bernoulli distribution $f_\theta(X) \equiv \, p_\theta(y=1 | X)$; (2) point-mass distributions with $p_{\bm{\theta}}(y|X) = \delta(y-f_{\bm{\theta}}(X))$; and (3) univariate regression when we model the output with a Gaussian distribution with a constant, but possibly unknown, variance, $p_{\bm{\theta}}(y|X) = \mathcal{N}(y|f_{\bm{\theta}}(X), \sigma_0^2)$.
Note that Wasserstein-1 distance does not depend on the observed label $y$ for any of these cases and that the definition of $f_\theta$ is circumstance-dependent.

\paragraph{The similarity regularization function $\bm{S}$:} The similarity regularization term encourages the discovery of a transformation that is significantly different from the identity transformation. A simple function to use is the inner product between $X$ and $T_{\phi}(X)$, which we will use to gain theoretical insights into our problem. Empirically, we choose the cosine similarity between $X$ and $T_{\mathbf{\phi}}(X)$, instead of the inner product, as it is a normalized measure and allows us to define a convenient threshold for similarity. Moreover, it does not depend on the magnitude of the transformations. For example, penalization with cosine similarity will not shrink the magnitude of the parameters of a linear transformation.
If we choose simple transformation functions $T_{\phi}$ such as linear functions, we can define even simpler similarity regularizers in terms of its parameters $\bm{\phi}$, too. An example is provided in the experiments in Section \ref{sec:nlp}.

Given the choices of $D$ and $S$, we define the optimization problem for Model INvariance Discovery (MIND) algorithm using the following transformation-agnostic loss function, obtained as the empirical expectation of Eq. (\ref{eq:subbed}):
\begin{equation}
    \min_{\bm{\phi}} \frac{1}{n} \sum_{i=1}^{n}\bigr [ | f_{\bm{\theta}}(X_i)) - f_{\bm{\theta}}(T_\phi(X_i))| + \lambda\cos{(X_i, T_\phi(X_i))}\bigr].
\label{eq:final}
\end{equation}

In the next subsections, we define different choices of $T_{\bm{\phi}}$ and show how to discover feature and pattern invariances.

\subsection{Identifying Transformation and Feature Invariances}
We use two choices for the transformation function $T_{\bm{\phi}}$ and show how we can discover transformation and feature invariances using them.

\paragraph{The residual transformation function $\bm{T}_{\bm{\phi}}$:} In clinical time series analysis, the features can be represented as $X = \{\mathbf{x}_t\in \mathbb{R}^d| t=1,\ldots, T \}$. Given the modeling power of residual blocks, we choose two residual network blocks with one-dimensional convolutional networks to model a generic $T_{\bm{\phi}}$. We will call this transformation ``\textit{Residual Transformation}'' and provide the details of its architecture in Appendix \ref{sec:models}. To interpret the learned transformation, we can compute the correlation coefficient between the original and transformed input data for each feature. In formal terms, for the $j^\mathrm{th}$ feature, we compute the Pearson correlation coefficients $\rho_j = \mathrm{corr}(x_j, T_{\bm{\phi}}(\mathbf{x})_j)$ over all timestamps and instances. Small values for $\rho_j$ mean the $j^\mathrm{th}$ feature can be transformed to a version that is less correlated with the original version, without a significant change in $p_{\theta}(y|X)$. When these small $\rho_j$ are found, we conclude that the model is insensitive to the $j^\mathrm{th}$ feature.

\paragraph{The gating transformation function $\bm{T}_{\bm{g}}$:} We also study simple affine transformations of the data, not only because we can use it to understand the sensitivities of the learning algorithms, but also we can obtain theoretical insights about the solution. The affine transformation, which we call the ``\textit{Gating Transformation}'', is described as $T_{\bm{g}, \bm{b}}(\mathbf{x}_t) = \mathbf{x}_t \odot\bm{g}+ \bm{b}$, where $\bm{g}\in [0, 1]^{d}, \bm{b}\in \mathbb{R}^d$, and $\odot$ denotes the element-wise product. We include the intercept $\bm{b}$ because it does not introduce any new invariance. This transformation performs a soft joint-variable selection on the input features. Each element $g_j$ for $j = 1, \ldots, d$ denotes the sensitivity of the model to the $j^\mathrm{th}$ feature in the time series. Since the elements of $\bm{g}$ are between 0 and 1, the interpretation of these coefficients is easier.  We call $g_j$ the \textit{MIND Score} of the corresponding feature $x_j$. 

\subsection{Identifying Pattern Invariances}
Predictive models on data such as time series and images do not rely solely on individual features; their decisions depend on more complex patterns in the data. Here we show how we use the gating transformation function to gain more insights into the invariance of the model to different patterns. To this end, we use an encoder-decoder pair to map the input to a hidden space, where each coordinate captures higher level patterns in the data. We use the gating transformation in the hidden space and decode the gated hidden representations using the decoder and pass it to the model.

The encoder-decoder pairs are usually obtained by training neural networks such as autoencoders. Because the reconstruction in autoencoders is not perfect, the analysis will be approximate. Here we define a lossless encoder-decoder pair for time series using orthonormal basis functions to analyze invariance to temporal patterns. 
We select a set of $K$ orthonormal temporal basis functions $\{a_k\}_{k=1}^{K}$ and map each dimension of the original time series to $K$ dimenstions: $x_j \mapsto (\langle x_j, a_1 \rangle a_1, \ldots, \langle x_j, a_K \rangle a_K)$, where $\langle \cdot, \cdot \rangle$ denotes the dot product over timestamps. The decoder for this encoder simply sums all the $K$ coordinates for each feature.
Thus, we combine the decoder and transformation functions and define the transformation $T_{\bm{g}, \bm{b}}$ for each feature to map from the $K$-dimensional space back to the $1$-dimensional input space, i.e., the transformation $T_{\bm{g}, \bm{b}}$ has $d$ linear maps each with a $K\times 1$ dimensional weight vector.

For analysis of the non-stationary temporal patterns in clinical time series, we use two sets of orthonormal basis functions: (1) Chebyshev polynomials of the first kind for analysis of a model's invariance to the mean, linear, quadratic, and the residual trends, and (2) pulse waves to analyze the impact of time of the events in the model. Figure \ref{fig:bases} in Appendix \ref{sec:basis} visualizes these basis functions.

%% file: analysis.tex
\section{Analysis of Discovering Feature Invariance}
In this section, we provide theoretical guarantees for discovery of invariances using the MIND scores. All results are small sample and non-asymptotic results. Throughout this section, to obtain uncluttered results, we use the inner product as the similarity regularization. We also assume that the data is normalized to have zero mean and finite variance.

First, we define a weaker version of feature invariance. 
\begin{definition}\textbf{Weak Invariance:} A differentiable multivariate function $f_{\bm{\theta}}(\mathbf{x})$ is weakly invariant to the $j^\mathrm{th}$ feature $x_j$, $j \in 1, \ldots, d$, if it satisfies $ \nicefrac{\partial f_{\bm{\theta}}(\mathbf{x})}{\partial x_j}  < C$ for all $\mathbf{x} \in \mathbb{X}$ and a small constant $C$. 
\end{definition}
The constant $C$ quantifies the degree of invariance, with smaller $C$ indicating more invariance. The assumption is weak, because it does not require the exact invariance $\nicefrac{\partial f_{\bm{\theta}}(\mathbf{x})}{\partial x_j}=0$. 

In the first result, we show the correctness of the algorithm in discovering the feature invariances.  If we call Eq.~\eqref{eq:final} the MIND equation when we specialize to the gating transformation, then we obtain the following:
\begin{theorem}
If $f_\theta$ is weakly invariant to the $j^\mathrm{th}$ feature with constant $C$, then the global solution for the MIND equation with $\lambda \geq C\nicefrac{\sum_{i=1}^{n}|x_{ji}|}{\sum_{i=1}^{n}x_{ji}^2}$ and the inner product regularization satisfies $g^*_j=0$. 
\label{thm:correctness}
\end{theorem}

The proof is based on showing that given the weak invariance assumption for the $j^\mathrm{th}$ feature, the partial derivative of the loss function with respect to $g_j$ is always positive. Thus the global minimum for the optimization problem occurs at the boundary $g^*_j=0$. The detailed proof is provided in Appendix \ref{sec:proofs}.

\paragraph{Remark.} Theorem \ref{thm:correctness} suggests that an ideal situation for use of MIND is to have a model that is invariant to most of its input variables. This observation can guide us in choosing the appropriate basis functions. 

To gain further insight into the structure of the solution, we study  the following simplified problem, which is designed with the goal of obtaining a closed-form solution, while being similar to the original problem. Given the homogeneous linear regression $y=\bm{\beta}^{\top}\mathbf{x}$, we look to find the closed-form solution of MIND scores $\bm{g}$ in terms of $\bm{\beta}$ and $\{\mathbf{x}_i\}_{i=1}^{n}$. 
Instead of the Wasserstein-1 loss considered in Eq.~\eqref{eq:final}, we further simplify the problem by solving the squared loss and inner product as follows:
\begin{equation}
    \bm{g}^* = \argmin_{\bm{g}}\frac{1}{n} \sum_{1}^{n} \left\{(\bm{\beta}^{\top}(\mathbf{x}_i-\bm{g}\odot\mathbf{x}_i))^2+\lambda(\bm{g}\odot\mathbf{x}_i)^{\top}\mathbf{x}_i\right\}
\label{eq:linear}
\end{equation}

\begin{theorem} Suppose the features have zero mean and the empirical correlation matrix $\bm{C}_n  = \frac{1}{n}\sum_{i=1}^{n}\mathbf{x}_i\mathbf{x}_i^{\top}$. The solution of Eq.~\eqref{eq:linear} for $\bm{g}$ is given as:
\begin{equation*}
    \bm{g}^* = \left[\bm{1}-\left(\lambda/2\right)(\bm{B}\bm{C}_n\bm{B})^{-1}\mathrm{diag}(\bm{C}_n)\right]_0^1,
\end{equation*}
where the diagonal matrix $\bm{B}$ is defined as $\bm{B} = \mathrm{diag}(\bm{\beta})$ and the clamp operator $[\cdot]_0^1$ is an element-wise projection to the interval $[0, 1]$.
\label{thm:simple}
\end{theorem}

The proof is based on finding the global solution of the quadratic function and projecting it to the feasible interval. To find the global optimum, we take the derivative with respect to $\bm{g}$ and set it to zero. The detailed proof is provided in Appendix \ref{sec:proofs}.

\paragraph{Remarks.} 
\begin{itemize}[topsep=0pt]
\item If the features are uncorrelated, i.e., $\bm{C}_n$ is diagonal, then $g_j = [1-\lambda \beta_j^{-2}/2]_0^1.$
\item If the features are uncorrelated and $\lambda \geq 2\beta_j^2$, then $g_j^* =0$. This shows how to control the sparsity of $\bm{g}^*$ via $\lambda$.
\item  The theorem implies that smaller (larger) values of the regression coefficient, $\bm{\beta}$, correspond to smaller (larger) values of MIND score, $\bm{g}^*$ . This observation helps us in interpreting and comparing the non-zero MIND scores $\bm{g}^*$.
\item The dependence of $\bm{g}^*$ on the empirical covariance matrix indicates that features that are correlated will share the MIND score with each other. 
\end{itemize}

%% file: exp.tex
We experiment on two types of healthcare data and two valuable prediction tasks defined on the publicly available MIMIC-III dataset \citep{johnson2016mimic}. First, we  predict the in-hospital mortality from multivariate clinical time series using the benchmark of \citet{harutyunyan2017multitask}. Next, we fine-tune the publicly available BioBERT \citep{lee2019biobert} on clinical notes to predict length of stay and analyze its invariant properties.

\begin{figure*}[t]
    \centering
    \begin{subfigure}[t]{0.45\textwidth}
    \centering
    \includegraphics[width=\textwidth]{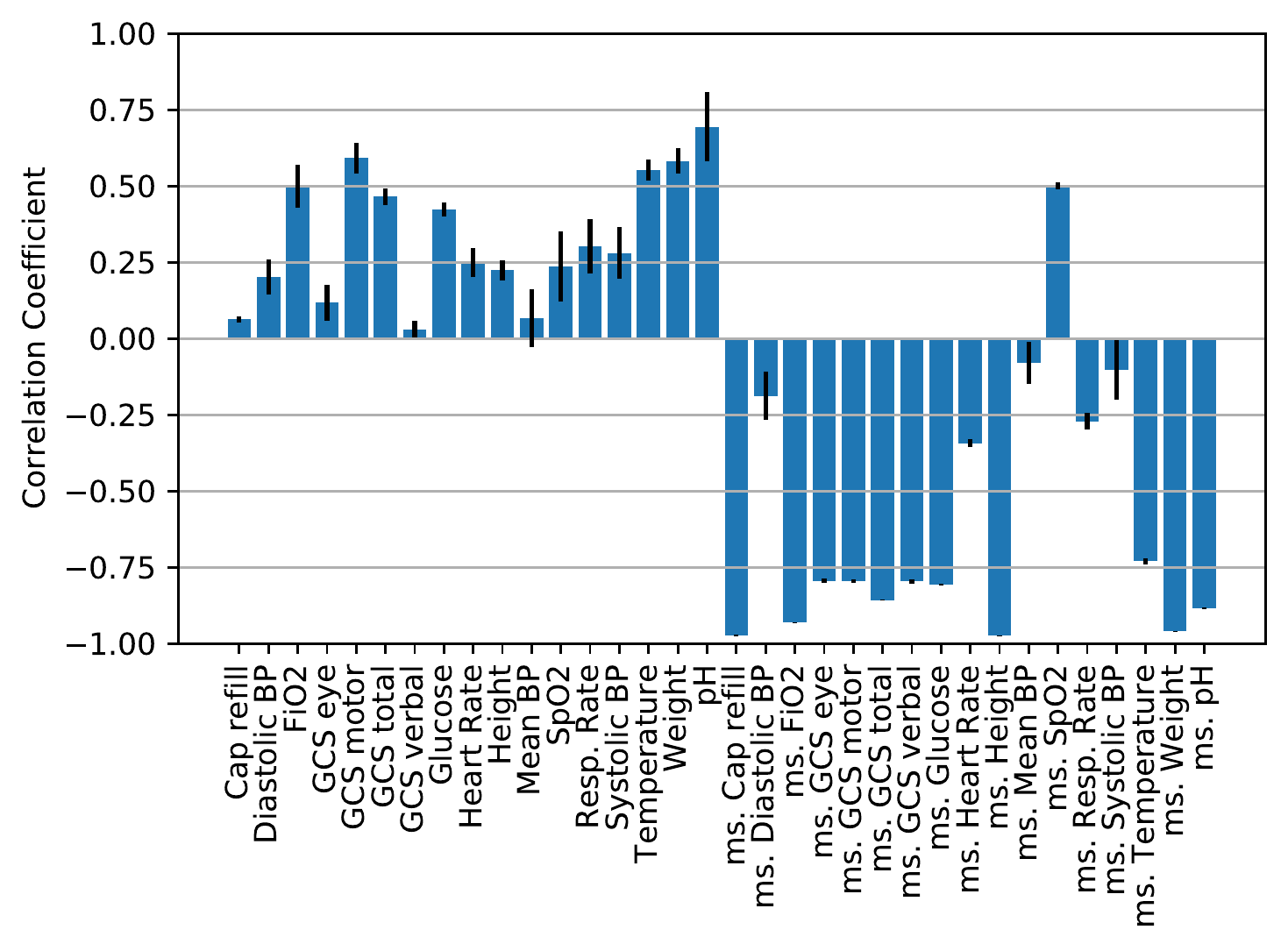}
    \caption{Residual Transformation}
    \label{fig:regular_residual}
    \end{subfigure}
    ~
    \begin{subfigure}[t]{0.45\textwidth}
    \centering
    \includegraphics[width=\textwidth]{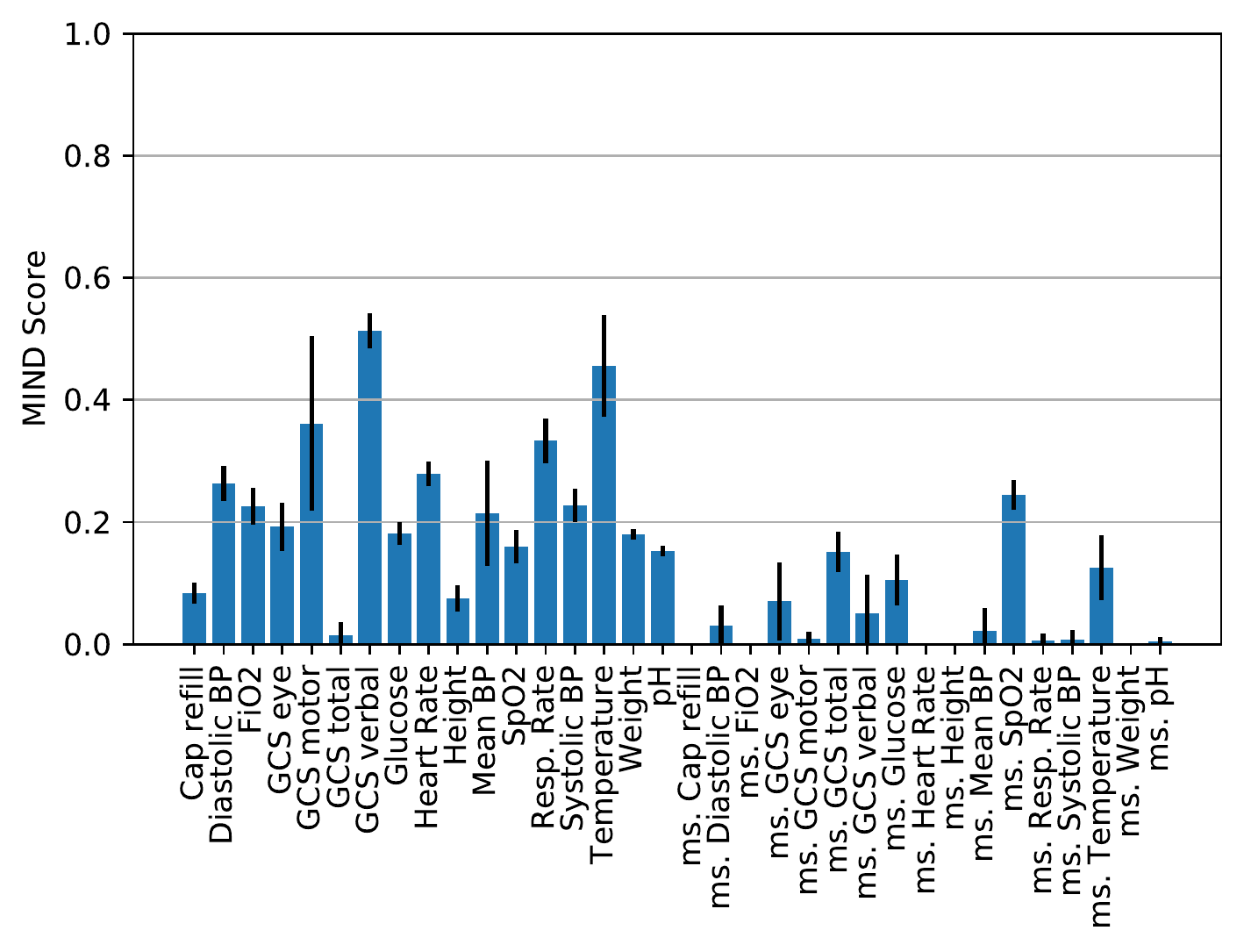}
    \caption{Gating Transformation}
    \label{fig:regular_gating}
    \end{subfigure}
    \caption{Discovering the invariance of the \textit{regularly-trained} LSTM model to the features using two types of transformation functions.   The missing value indicators are identified with ``ms.'' at the beginning.  
    }
    \label{fig:regular}
\end{figure*}

\begin{figure*}
    \centering
    \begin{subfigure}[t]{0.45\textwidth}
    \centering
    \includegraphics[width=\textwidth]{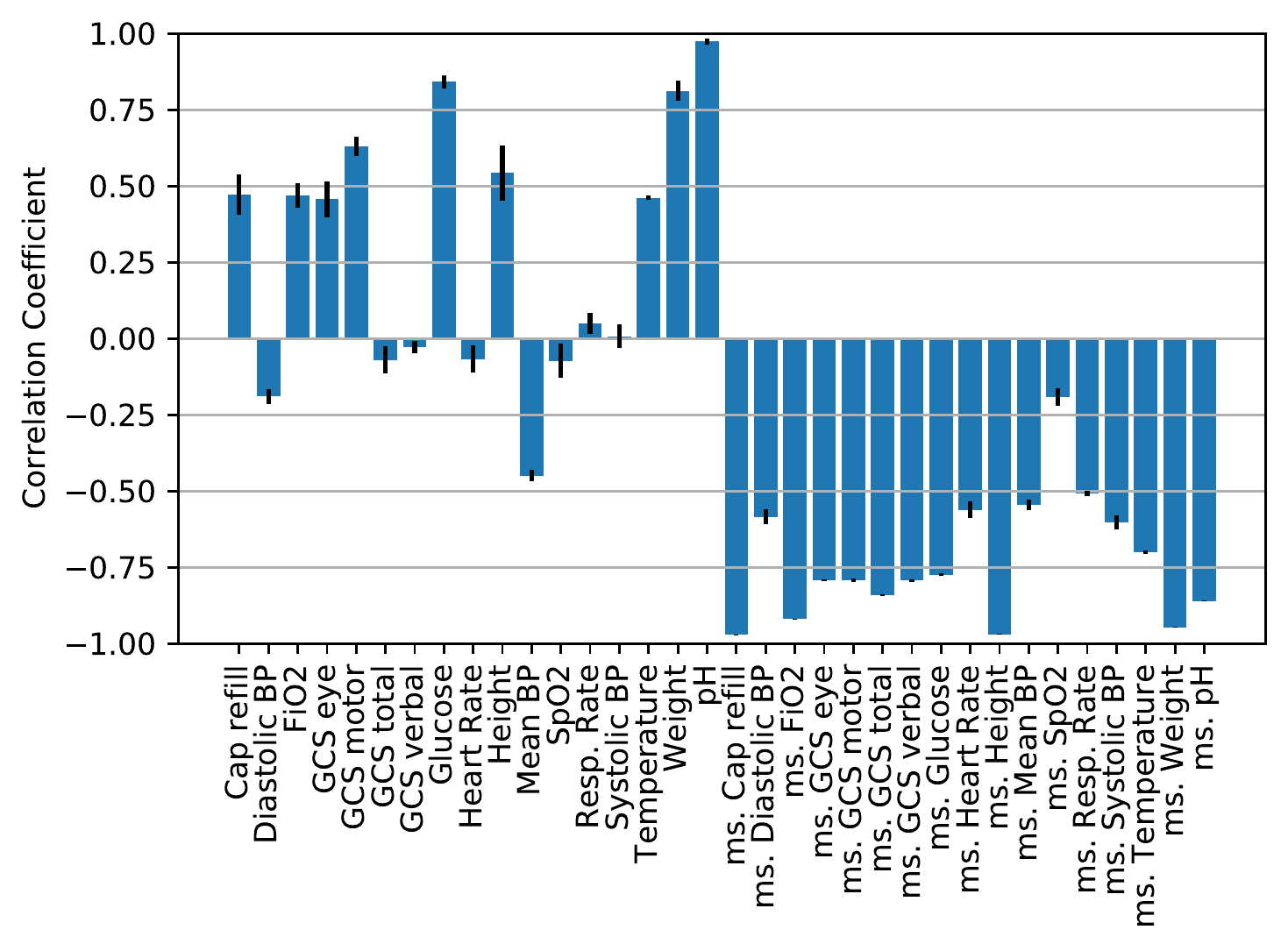}
    \caption{Residual Transformation}
    \label{fig:adversarial_residual}
    \end{subfigure}
    ~
    \begin{subfigure}[t]{0.45\textwidth}
    \centering
    \includegraphics[width=\textwidth]{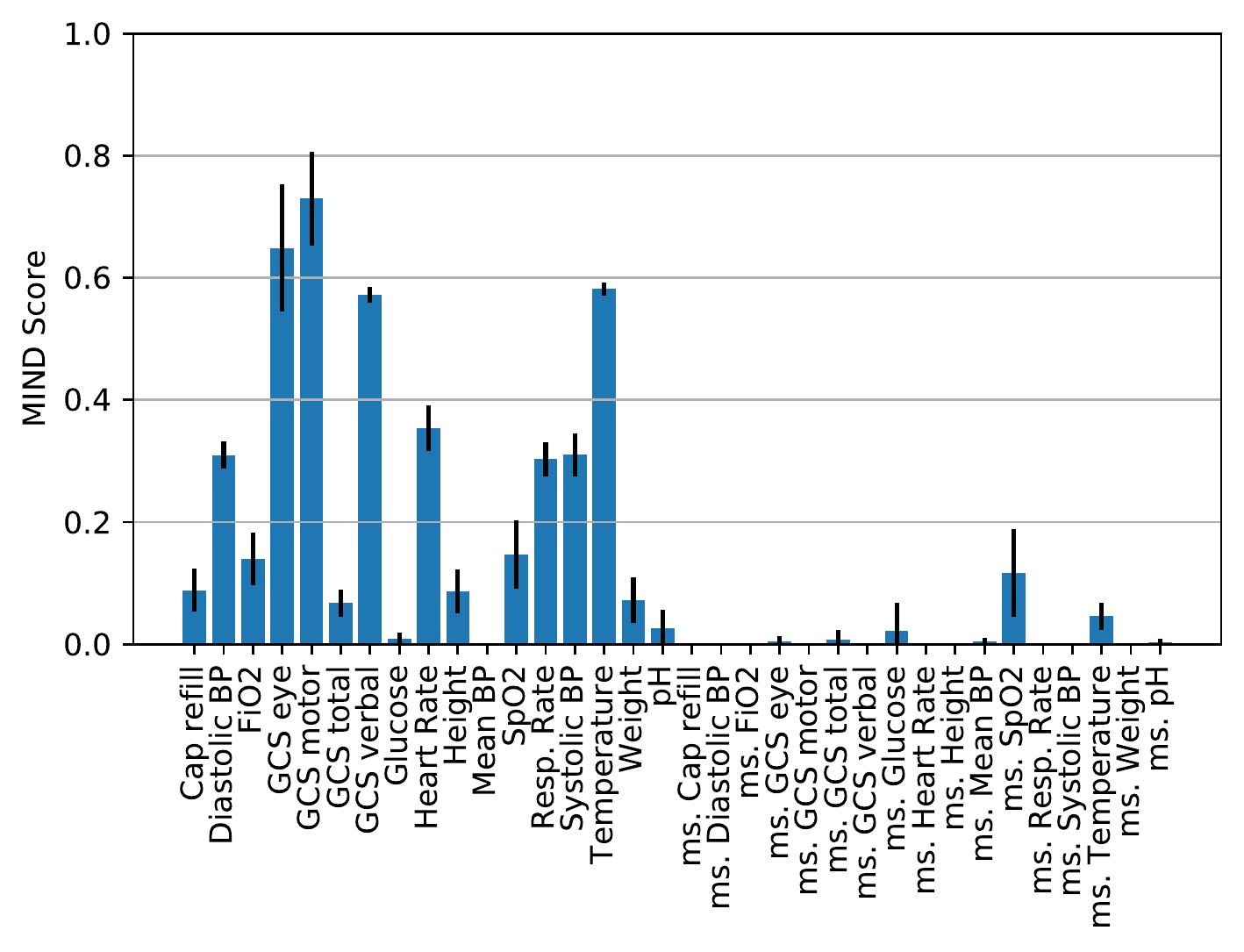}
    \caption{Gating Transformation}
    \label{fig:adversarial_gating}
    \end{subfigure}
    \caption{Discovering the invariance of the \textit{adversarially-trained} LSTM model to the features using two types of transformation functions. The adversarially trained model is invariant to more features, including many missing value indicators (identified with ``ms.''.).  
    }
    \label{fig:adversarial}
\end{figure*}

\subsection{Experiments with clinical multivariate time series}
\subsubsection{Data, Training, and Baseline Details}
\label{sec:training_details}
\paragraph{Dataset and tasks:} We evaluate the proposed algorithm on the in-hospital mortality  benchmark task \citep{harutyunyan2017multitask}, using the publicly available MIMIC-III dataset \citep{johnson2016mimic}. The objective of this task is to predict mortality outcome based on a time series of multivariate observational data from patients in the intensive care unit (ICU). This is one of the most commonly studied tasks in healthcare analytics and is widely used in  estimating patients' clinical risk and managing costs in hospitals \citep{choi2016retain,rajkomar2018scalable}.

We extract the patient sequences from the MIMIC-III database 
and partition the data into training and testing sets, following \citet{harutyunyan2017multitask}.
To make the interpretation of the results easier, we treat the ordinal values as real values and represent each ordinal variable with a single real variable. We normalize the features as described by \citet{bahadori2019temporal}. 

After preprocessing, the features are in the form of multivariate time series of length 60 timestamps. We have 17 features (listed in Table \ref{tab:names} in the appendix) and we add 17 more binary variables that indicate a missing measurement for each of the features. Following \citet{lipton2016learning}, we substitute the missing values with zeros. Thus, the input features are $34\times60$ time series and the labels are binary.

We hold out $15\%$ of the training data as a validation set 
for tuning the hidden layer sizes and hyperparameters. 
We report the test results based on the best validation performance.
For optimization, we use Adam \citep{kingma2014adam}
with the AMSGrad modification \citep{reddi2018convergence} with batch size of $100$. 
We halve the learning rate after plateauing for $10$ epochs (determined on validation data)
and stop training after the learning rate drops below $5\times 10^{-6}$.

\paragraph{Regular and adversarial training of the base predictor $f_{\bm{\theta}}$:} We choose to train an LSTM network because this is a common benchmark for analysis of clinical time series \citep{lipton2016learning}. We use a two-layered LSTM with 200 hidden neurons in each layer. Given the relationship between adversarial training and invariances \cite{jacobsen2019excessive}, we train the model both in a regular (non-adversarial) and adversarial settings. For adversarial training we use the projected gradient descent algorithm \citep{madry2018towards}. Our goal of using adversarial training is to avoid the possible rediscovery of easy-to-avoid adversarial examples in the transformation $T_\phi$ found by our algorithm. The regular and adversarially trained models achieve test AUC of 0.8600 and 0.8554, respectively, which are comparable to the results reported by \citet{bahadori2019temporal}. 

For comparison, we also train a second network based on the transformer encoder \citep{vaswani2017attention} and convolutional layers (details provided in Appendix \ref{sec:models}). The regular and adversarially-trained transformer-based models achieve test AUC of 0.8539 and 0.8533, respectively. Given the superior performance of LSTM, we present the main experiments only with LSTM and use the transformers only in the temporal patterns analysis for comparison.

\paragraph{Training of the transformation $T_{\bm{\phi}}$:} Fixing the parameters of the trained LSTM predictor $f_{\bm{\theta}}$ as above, we train both the gating and residual transformations, defined above, using Eq. (\ref{eq:final}). Further implementation details are in the appendix. For the gating transformation, we initialize the transformation matrix as an identity matrix whose elements are perturbed by independent Gaussian noise $\mathcal{N}(0, 0.02)$. We select a value for $\lambda$ and the weight decay to find solutions that satisfy two upper limits on maximum $W_1$ distance (0.05) and cosine similarity (0.5). To capture the training variations because of the random initialization, we repeat the training 24 times each with different random initialization. Then, we pick the top five best trained models based on the validation loss and report the mean and standard deviation.

\paragraph{Baselines:} To the best of our knowledge, there is no appropriate baseline to compare the invariance discovery performance. However, we can construct baselines using the popular feature attribution techniques such as saliency maps \citep{simonyan2013deep}, Integrated Gradients \citep{sundararajan2017axiomatic}, and SHAP \citep{lundberg2017unified}. We randomly choose 100  patients and find the attribution maps using each of the above methods via their implementation in Captum \citep{captum2019github}. We find the average of the absolute value of the maps and compare them with our MIND scores in the sanity check scenarios introduced in \citet{adebayo2018sanity}.

\begin{figure*}[h]
    \centering
    \begin{subfigure}[t]{0.45\textwidth}
    \centering
    \includegraphics[width=\textwidth]{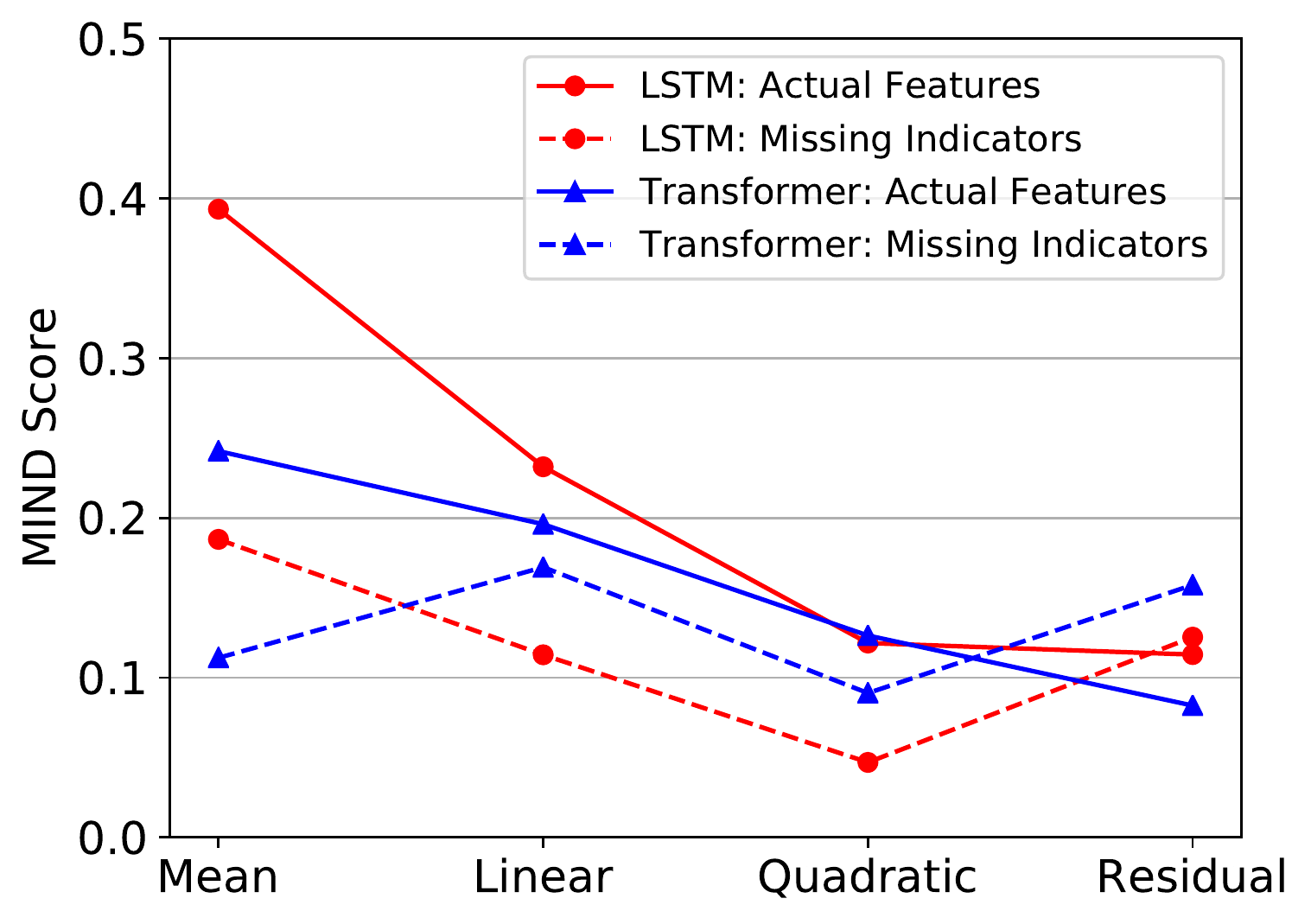}
    \caption{Chebyshev basis}
    \label{fig:cheby_result}
    \end{subfigure}
    \begin{subfigure}[t]{0.45\textwidth}
    \centering
    \includegraphics[width=\textwidth]{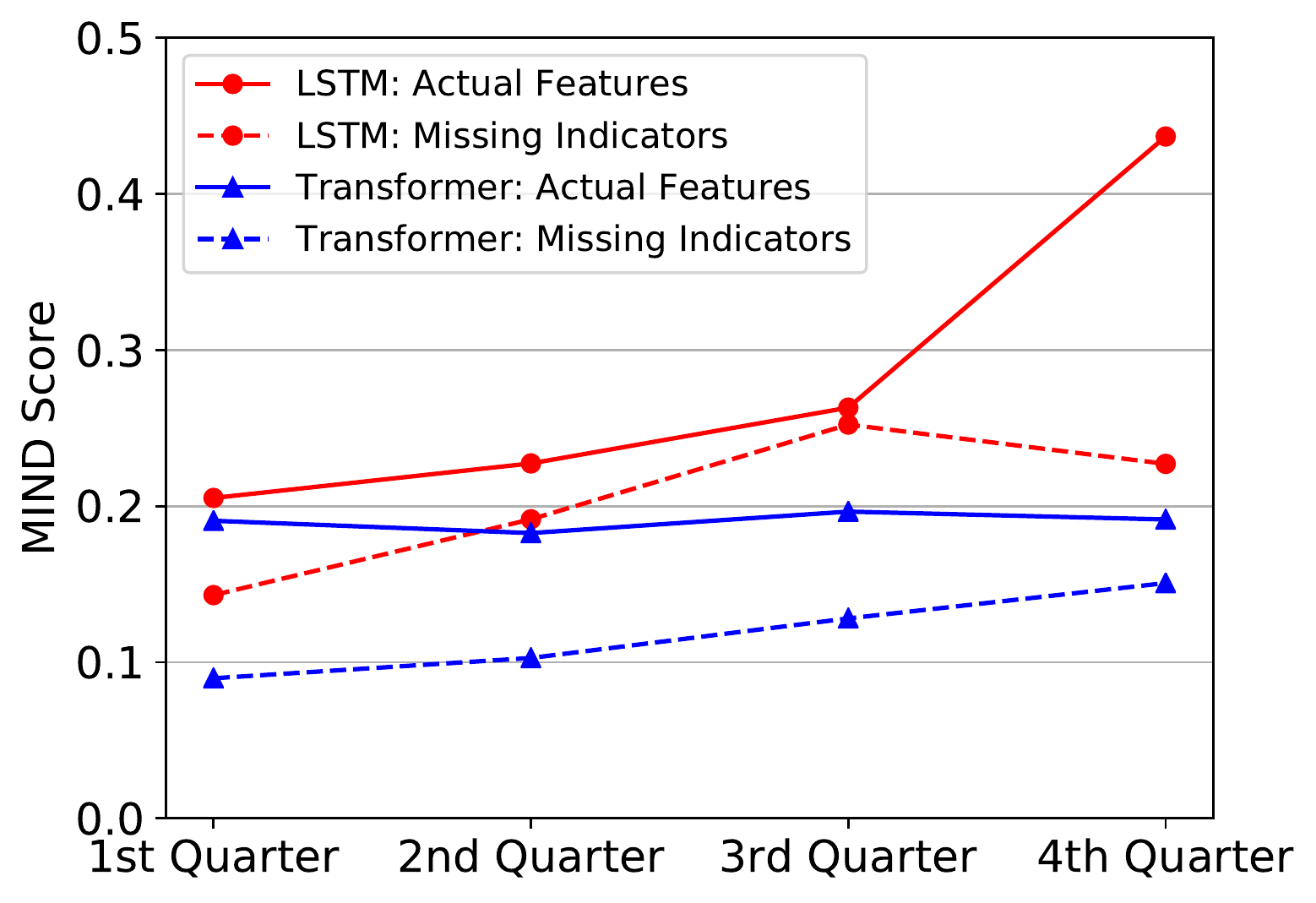}
    \caption{Pulse basis}
    \label{fig:pulse_result}
    \end{subfigure}
    \caption{Identifying the influential temporal trends in the \textit{regularly-trained} networks' decisions. The numbers are the average over all features.  (\subref{fig:cheby_result}) We decompose the time series to mean, linear, quadratic, and residual trends. The plot shows that the mean and linear trends in the data play a bigger role in the networks' decisions.
    (\subref{fig:pulse_result}) Using the pulse basis (Figure \ref{fig:pulse_bases}), we evaluate the invariance of the model to the information in four time periods of the time series. Note that overall more recent information, especially in the actual features, play a bigger role in the LSTM's decisions compared to the Transformer-based network. }
\end{figure*}

\subsubsection{Results and Analysis}

\paragraph{MIND Scores for LSTM.} Figure \ref{fig:regular}  shows the correlation coefficients and MIND scores obtained by the residual and gating transformations on the regularly trained LSTM model.  The full name of the features in these plots are provided in Table \ref{tab:names} in Appendix \ref{sec:names}. Using the residual transformation function, the average correlation coefficients for the actual features and the missing value indicators are $0.3281$ and $-0.5877$, respectively. Similarly, the average MIND scores for the actual features and the missing value indicators are $0.2301$ and $0.0485$, respectively. While this shows the significance of the main features, it also underlines the contributions of the missing value indicators. 

Looking at the main features we observe that the one of the coma scores has the smallest score, because the total GCS score can be explained by the other GCS scores, as shown in the correlation matrix in Figure \ref{fig:corr} in Appendix \ref{sec:models}. The MIND score for the ``height'' feature is also small, indicating the minor role of patients' height in the mortality in the intensive care units. The ``capillary refill rate'' also receives small scores with both methods. Examining the data shows that capillary refill rate is available only for 0.28\% of the timestamps; and the learned transformation shows that the feature is not reliable in prediction of the outcome for a large number of patients, as might be expected from its low incidence rate. It is important to note that we have not explicitly trained the LSTM with any regularization criteria that enforce sparsity or variable selection.

The results show that the missing value indicator for oxygen saturation $\mathrm{SpO}_2$ is exceptionally important. Clinically, 
the absence of $\mathrm{SpO}_2$ may not inform providers if the patient is receiving enough oxygen to prevent hypoxemia and tissue hypoxia. The result might also be due to the label leakage in the benchmark cohort construction, as suggested by \citet{wang2019mimic}.
We further observe that the mean arterial blood pressure (MAP) is less important compared to the systolic (SBP) and diastolic (DBP) pressures.  While the systolic and diastolic pressure measurements are ubiquitous, a direct measurement of the MAP is invasive; and it can also typically be estimated as $\mathrm{MAP}=(2\mathrm{DBP} - \mathrm{SBP})/3$, which is a common clinical rule-of-thumb \cite{brzezinski}.

\paragraph{The impact of adversarial training.} Figure \ref{fig:adversarial} shows the correlation coefficients and  MIND scores on the adversarially trained LSTM. To have a fair comparison between the two models, we ensure that the Wasserstein loss term for both the regular and adversarially trained models are equal to $0.05$. In the adversarially trained model, the average correlation coefficients for the actual features and missing value indicators are $0.2845$ and $-0.7270$, respectively.
Similarly, the average MIND scores for the actual features and missing value indicators are $0.2615$ and $0.0120$, respectively. The results show that, by using adversarial training, the model no longer depends on many of the missing value indicators.  Notice that the sensitivity scores for features such as ``GCS total'' have further decreased in comparison to Fig.~\ref{fig:regular}. This experiment is in line with the finding of \citet{jacobsen2019exploiting}, who argue that the perturbation-based robustness can create new invariances in the model. 

Figures \ref{fig:regular} and \ref{fig:adversarial} show that the results with gating and residual transformations are quite similar. Given the computational efficiency of the gating transformation, we present the next set of experiments only with the gating transformation.

\paragraph{The impact of trends.} To analyze the contribution of non-stationary trends in time, we decompose each individual time series using the first three Chebyshev polynomials (Fig.~\ref{fig:cheby_bases}), combined with the residual value between the time series and these three basis elements. Figure \ref{fig:cheby_result} shows the MIND scores for each component, averaged over all features. It shows that the lower order trend of the time series play the biggest role in the LSTM's decisions. A similar trend can be seen in the Transformer-based model. This observation justifies the findings by \citet{lipton2016learning} that hand-engineered features based on trends work well for clinical time series classification tasks.

\paragraph{The impact of temporality.} Using the pulse basis in Fig.~\ref{fig:pulse_bases}, we evaluate the invariance of the model to the information in four distinct time periods of the time series. Again, we average the MIND scores over all features. Figure~\ref{fig:pulse_result} shows that overall more recent information, especially in the actual features, play a bigger role in the LSTM's decisions. This can be due to either physiological reasons or an LSTM's structural  sensitivity to more recent inputs. In contrast to the LSTM, the Transformer-based model pays equal attention to all time periods.

 \begin{table}[t]
    \centering
    \caption{Spearman rank correlation between baselines and our coefficients: correlation (p-value).}
    \small
    \begin{tabular}{@{}l|c|c|c@{}}
     & \textbf{Int. Grads} & \textbf{SHAP} & \textbf{Saliency}\\
     \midrule
     \textbf{LSTM} & 0.438 (0.010) & 0.441 (0.009) & 0.723 (1.35e-06)\\
     \textbf{Trans.} & 0.110 (0.535) & 0.162 (0.357) & 0.859 (7.90e-11)\\
     \textbf{CNN} & 0.076 (0.668) & -0.031 (0.862) & -0.050 (0.777) \\
    \end{tabular}
    \label{tab:corr_baseline}
\end{table}
\subsubsection{Comparison to Baselines and Sanity Check}
\label{sec:sanity}
\citet{adebayo2018sanity} argue that any model explanation tool should be sensitive to randomizations in the model and data. Because MIND  characterizes the behavior of a model on the entire dataset, data randomization tests cannot be used. Thus, in this section, we shuffle the weights independently in each layer of the target neural network and find the correlation between the invariances of the model with a shuffled layer and the invariances of the original model.

\begin{figure}[t]
    \centering
    \includegraphics[scale=0.5]{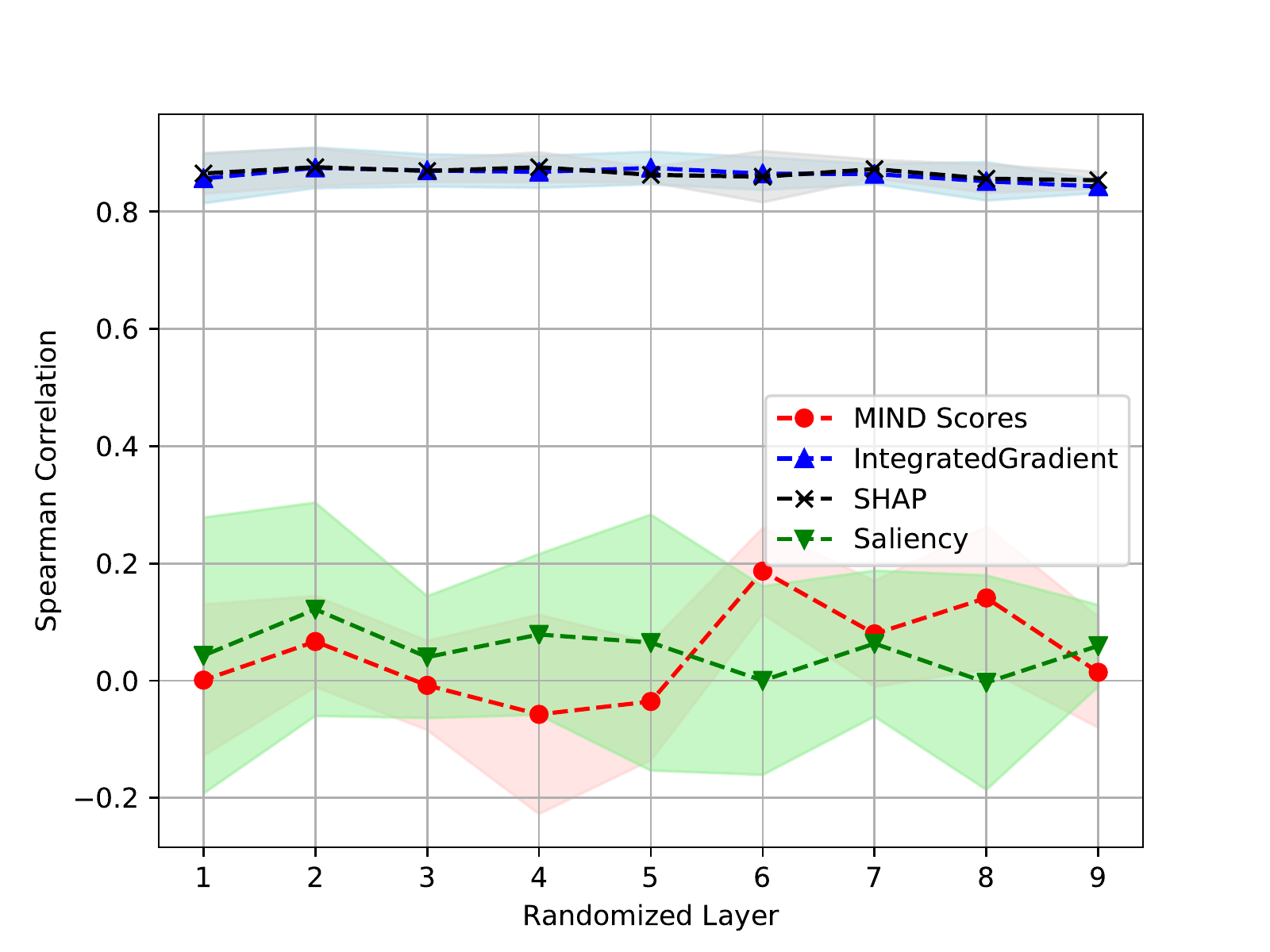}
    \caption{Sanity check (\textit{lower the better}): Independent randomization of each layer and the correlation between the MIND scores on the randomized and original models.}
    \label{fig:randomization}
\end{figure}

Because shuffling layer weights might destabilize LSTMs and Transformers, we use an 8-layer CNN model with a dense prediction layer on top. The detail of the model is given in Appendix \ref{sec:models}. Because the CNN model's test AUC is $0.8426$, subpar to our LSTM and Transformer-based model, we did not include it in the previous experiments. We create 5 instances of weight-shuffled models for each of the 9 layers and report the mean and standard deviation over 5 instances.  The results in Figure \ref{fig:randomization} show that the MIND score is quite sensitive to the randomization in the model and passes the sanity checks.

Table \ref{tab:corr_baseline} shows the Spearman rank correlation between the MIND score and the average of absolute values of the explanations by three baseline techniques on three models. The results show a wide variation across methods and models: there is a strong (but not perfect) correlation between the average saliency maps with the MIND scores on LSTMs and Transformers, but the correlation becomes insigificant on the CNN model. Thus, the results confirm that the invariance map by MIND scores cannot be simply calculated by finding the average attribution map by any of these attribution techniques.

\begin{table}[t]
    \caption{Examples of tokens that received high and low MIND scores in our analysis, presented in (token, MIND score) format.}
    \label{tab:biobert_examples}
    \centering
    \begin{tabular}{p{7.5cm}}
    \toprule
    \textbf{High MIND Score}\\
    \midrule
(premature, 0.969)
(twin, 0.881)
(infant, 0.877)
(with, 0.786)
(tube, 0.784)
(respiratory, 0.779)
(no, 0.760)
(membrane, 0.754)
(surf, 0.747)
(week, 0.747)
(lungs, 0.743)
(consolidation, 0.741)
(baby, 0.737)
(pre, 0.717)
(diagnosis, 0.716)
(for, 0.712)
(and, 0.707)
(mother, 0.704)
(newborn, 0.701)
(pneumonia, 0.696)
(weeks, 0.688)
(valve, 0.686)
(bilateral, 0.672)
(normal, 0.669)
(feeds, 0.668)
(distress, 0.661)
(failure, 0.659)
(triple, 0.652)
(lung, 0.635)
(pregnancy, 0.629)  \\
\midrule
\textbf{Low MIND Score}\\
\midrule
(compared, 0.037)
(aware, 0.054)
(shadow, 0.061)
(cost, 0.063)
(cavity, 0.066)
(bases, 0.067)
(apex, 0.069)
(added, 0.069)
(upright, 0.070)
(obtained, 0.070)
(again, 0.070)
(flutter, 0.070)
(reviewed, 0.071)
(moderately, 0.071)
(basal, 0.071)
(preserved, 0.072)
(arch, 0.073)
(able, 0.073)
(daily, 0.073)
(please, 0.074)
(add, 0.074)
(partial, 0.075)
(hi, 0.075)
(factors, 0.075)
(pat, 0.075)
(telephone, 0.076)
(access, 0.076)
(mouth, 0.076)
(imaging, 0.076)
(acoustic, 0.076)\\
\bottomrule
    \end{tabular}
\end{table}

\subsection{Experiments on Clinical Notes}
\label{sec:nlp}
In the second set of experiments, we use clinical notes in the MIMIC-III dataset to predict the length of stay (LoS) at hospital admission. LoS is one of the canonical predictive modeling tasks in healthcare; for example the LACE clinical risk scores  \citep{ben2012simplified,van2012lace+,van2010derivation,gruneir2011unplanned} require an estimate of the length of stay of a patient soon after they are admitted to the hospital.

\subsubsection{Experiment Setup}
\paragraph{Cohort selection.}
We create a cohort of 49,212 patients who have at least one note in the first day of visit. We also exclude all patients whose length of stay is less than a day to avoid misleading results due to reassignment. We preprocess the notes by a publicly available script\footnote{\scriptsize\url{https://github.com/sudarshan85/mimic3-utils/}\normalsize}. We concatenate all the notes that a patient receives during their first day. The LoS in our dataset measures as the number of hours since the first admission. We randomly hold out 15\% of the data for validation purposes.

\paragraph{Base Model and training.} We fine-tune the publicly available BioBERT \citep{lee2019biobert} model according to the recipe by \citet{Wolf2019HuggingFacesTS}\footnote{\scriptsize \url{https://github.com/huggingface/transformers/blob/master/examples/run_glue.py}\normalsize}. We also use BioBERT's tokenizer, which has 28,996 tokens. During training and validation, we set the maximum number of tokens in each document is set to 512. We use the $L_1$ loss function for robust regression purposes. 

\paragraph{Invariance discovery.} We use the gating function as the transformation to find MIND scores for each words in the vocabulary. We implement the gating function using a token-specific scalar (embedding to a 1-dimensional space) and multiply the token-specific scalar to each dimension of the embedded sequence. For the similarity regularizer, we use an $L_1$ regularization on the weights of the 1-dimensional embedding matrix with $\lambda=0.1$. The rest of the training details is the same as the procedure described in Section \ref{sec:training_details}. We stop the training when the Wasserstein term in the validation loss becomes $W_1 = 0.08$ hours.
\subsubsection{Results and Analysis}
We calculate the MIND scores for all tokens and compare the value of all tokens to the subset of the tokens that are listed in the Merriam Webster Medical Dictionary \citep{merriam1995merriam}\footnote{Using a medical dictionary is our best effort, but it is only an approximate way for classifying tokens into medical and non-medical categories. Many words can have several medical and non-medical uses.}. The average MIND score for all tokens and medical tokens are $0.117$ and $0.140$, respectively. Using the two-sample Kolmogorov-Smirnov test, we confirm that these two populations of MIND scores have significantly different distributions with test statistic of $0.146$ and p-value of $4.25\times 10^{-63}$. 

Several examples of tokens with high and low MIND scores are listed in Table \ref{tab:biobert_examples}. Inspecting the tokens, we see meaningful tokens such as ``premature'', ``pneumonia'', and ``failure'' in the high MIND score list. In the low MIND score list, we see neutral words such as ``compared'', ``added'', and ``please''. These examples show how we can investigate the model quality and ensure that our model uses the right tokens for prediction.

\begin{figure}[t]
    \centering
    \includegraphics[scale=0.5]{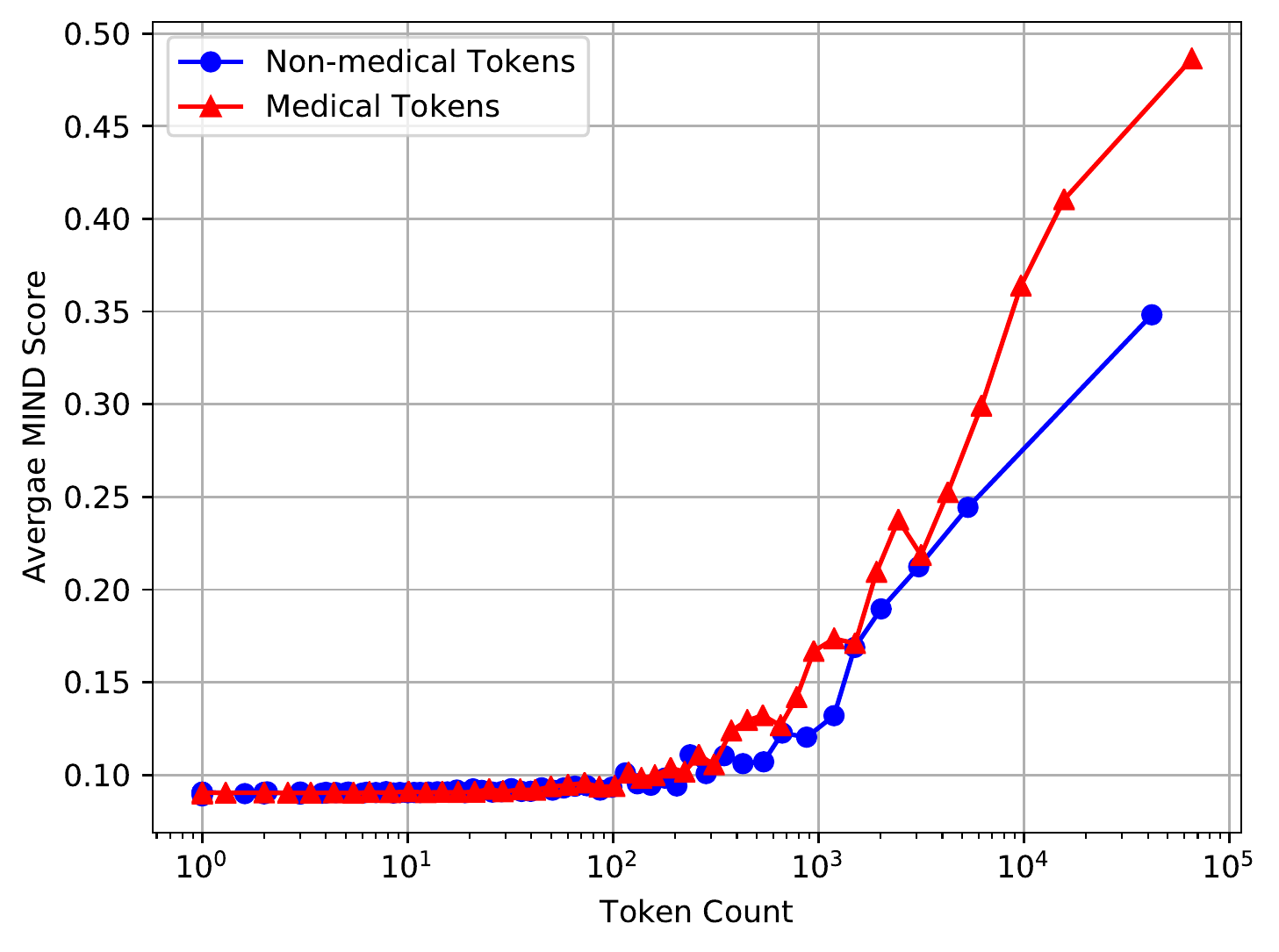}
    \caption{The impact of token frequency on the MIND score. The x-axis is in the logarithmic scale, in the bins of 100 data points.}
    \label{fig:frquency_bert}
\end{figure}

Using the MIND scores, we observe that BioBERT relies mostly on the most frequent tokens for analysis of text. In Figure \ref{fig:frquency_bert} we plot the relationship between MIND scores and the token counts in the training dataset. To show a smoother plot, we show the averages in bin sizes of 100. A further interesting observation is that the Spearman rank correlation between frequency and MIND scores is high for medical vs. non-medical tokens: $0.641$ vs $0.419$. This can be because the medical token list does not include many neutral but very frequent tokens that receive low MIND scores.

%% file: conc.tex
In this work, we proposed the MIND framework for discovering invariances of healthcare models. We analyzed its correctness and the structure of its solution. Using experiments on two key healthcare analytics tasks and two different data types, we demonstrated how we can use MIND scores in practice. In particular, we used the MIND scores to evaluate the efficacy of the indicator approach for missing value treatment strategy and study the impact of adversarial training.

We can generalize the MIND algorithm to discover the invariances in the data too. One approach can be not fixing the target model and adding the MIND loss terms to the log-likelihood loss function used for the training of the model. Finding invariances in the data can help us in data augmentation and improvement of the accuracy of predictive models. We will explore this topic in future works.

%% file: basis.tex
\begin{figure}
\vspace{-0.1in}
    \centering
    \begin{subfigure}[t]{0.45\textwidth}
    \centering
    \includegraphics[width=\textwidth]{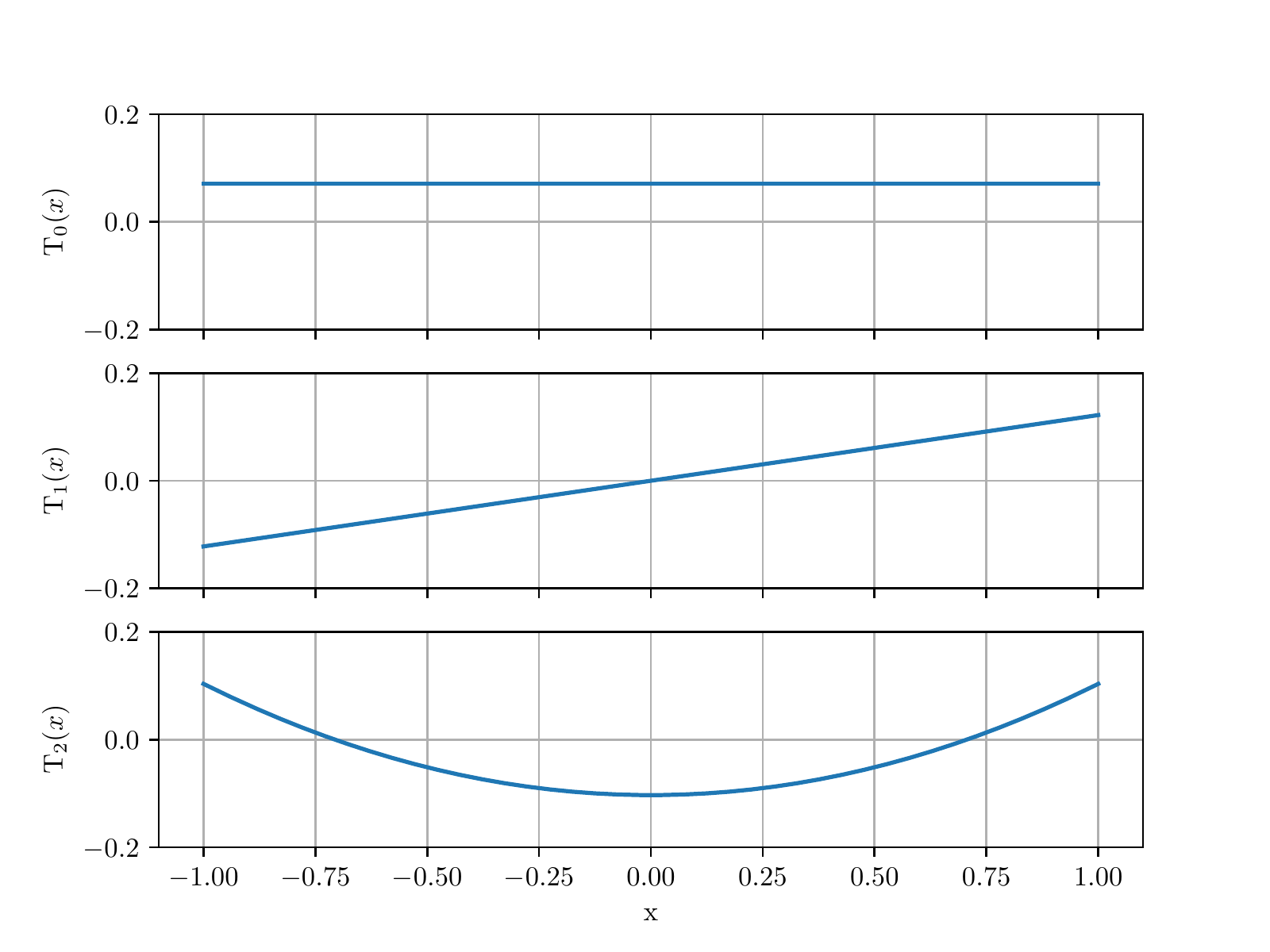}
    \caption{Chebyshev bases of first kind}
    \label{fig:cheby_bases}
    \end{subfigure}
    \begin{subfigure}[t]{0.45\textwidth}
    \centering
    \includegraphics[width=\textwidth]{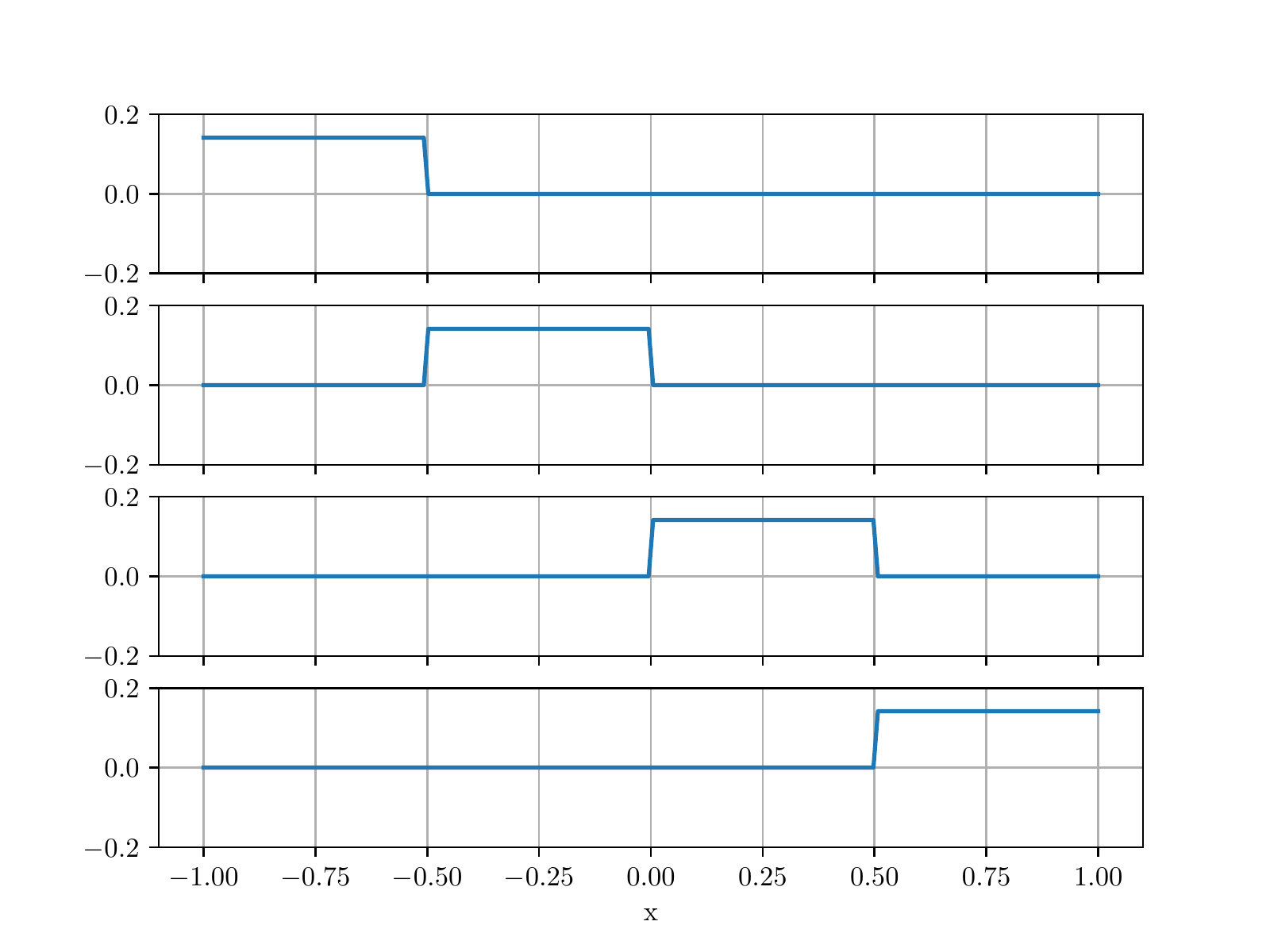}
    \caption{Pulse basis functions}
    \label{fig:pulse_bases}
    \end{subfigure}
    \caption{(\subref{fig:cheby_bases})  The first three Chebyshev polynomials of first kind, capturing the mean, linear, and quadratic trends in the data. We also use the residual of the time series as the fourth dimension.
    (\subref{fig:pulse_bases}) The pulse basis functions.)}
\label{fig:bases}
\vspace{-0.1in}
\end{figure}

%% file: proofs.tex
\begin{figure}[t]
\vspace{-0.1in}
    \centering
    \includegraphics[scale=0.6]{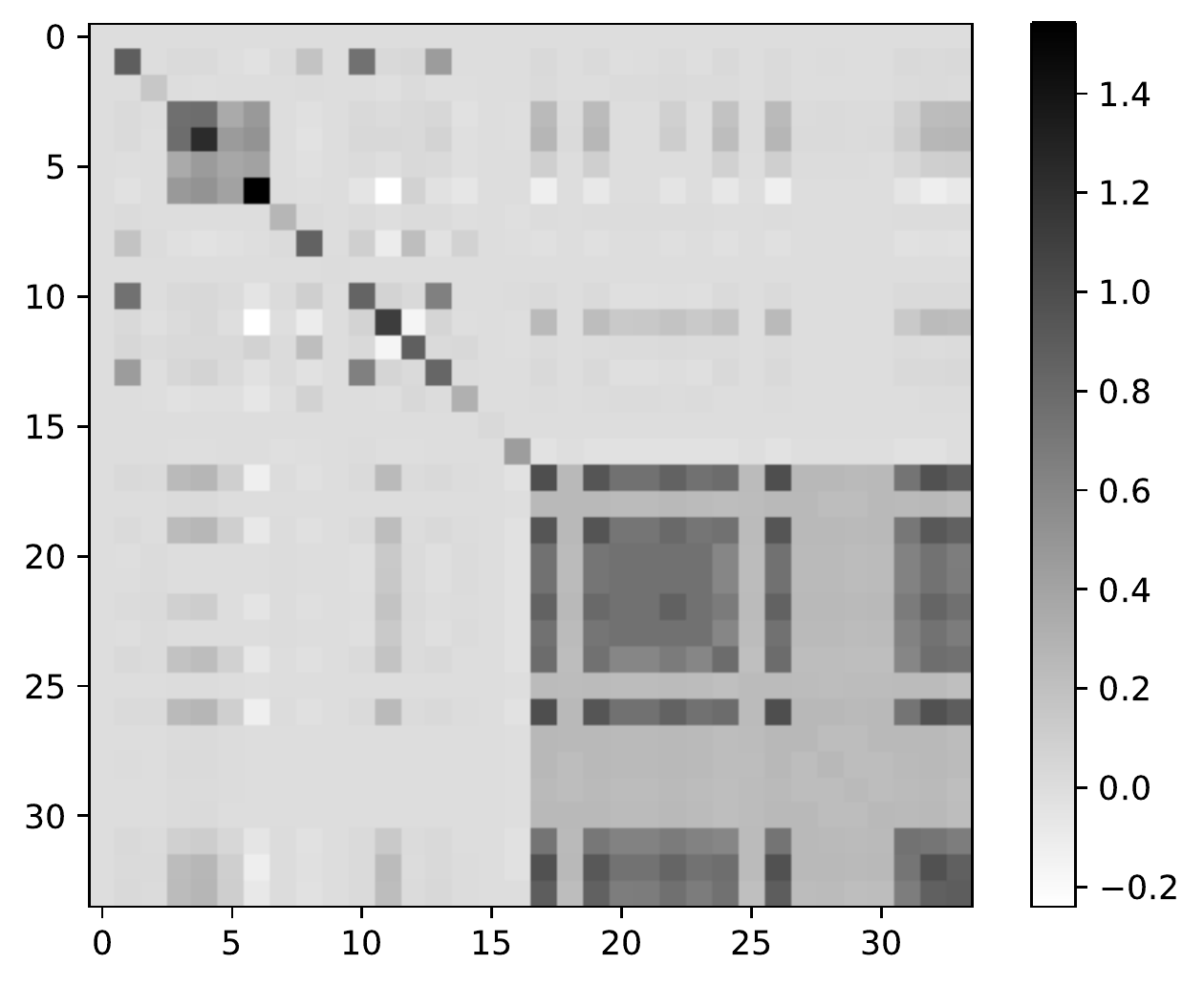}
    \caption{The empirical correlation matrix of the training data.}
\vspace{-0.1in}
    \label{fig:corr}
\vspace{-0.1in}
\end{figure}

\subsection{Theorem \ref{thm:correctness}}
\begin{proof} 
We start by lower-bounding the partial derivative of the loss function:
\begin{align*}
&\frac{\partial}{\partial{g_j}}\left(|f_{\bm{\theta}}(\mathbf{x}) - f_{\bm{\theta}}(\bm{g}\odot \mathbf{x})|+\lambda (\bm{g}\odot\mathbf{x})^{\top}\mathbf{x} \right)\\
    &  = -x_j\mathrm{sgn}(f_{\bm{\theta}}(\mathbf{x}) - f_{\bm{\theta}}(\bm{g}\odot \mathbf{x})) \left.\frac{\partial}{\partial{x_j}}f_{\bm{\theta}}(\mathbf{x})\right|_{\mathbf{x} = \bm{g}\odot\mathbf{x}} + \lambda x_j^2\\
    & \geq -|x_j| \left|\left.\frac{\partial}{\partial{x_j}}f_{\bm{\theta}}(\mathbf{x})\right|_{\mathbf{x} = \bm{g}\odot\mathbf{x}} \right|+ \lambda x_j^2. \\
    & > -C|x_j| + \lambda x_j^2,
\end{align*}
where in the last step we have used the weak invariance assumption that $\left|\frac{\partial}{\partial{x_j}}f_{\bm{\theta}}(\mathbf{x}) \right| < C$ for all $\mathbf{x}$ and a constant $C$. For clarity, we use $\left. h(\mathbf{x})\right|_{\mathbf{x}_0}$ to denote the value of a function $h(\mathbf{x})$  at a point $\mathbf{x}_0$. Taking empirical expectation, and choosing $\lambda = C\nicefrac{\sum_{i=1}^{n}|x_{ji}|}{\sum_{i=1}^{n}x_{ji}^2}$, the partial derivative becomes always positive. This means the loss function monotonically increases in terms of $g_j$. Thus, the global minimum of the gradient descent occurs at the boundary $g_j=0$.
\end{proof}

\subsection{Theorem \ref{thm:simple}}
\begin{proof} We note that the unconstrained objective function is a quadratic function in terms of $\bm{g}$. Thus, we can find the solution to the constrained objective function by finding the global solution of the quadratic function and projecting it to the feasible interval. To find the global optimum, we take the derivative with respect to $\bm{g}$ and set it to zero.
\begin{align*}
    \nabla_{\bm{g}}\frac{1}{n} \sum_{1}^{n}& \left\{(\bm{\beta}^{\top}(\mathbf{x}_i-\bm{g}\odot\mathbf{x}_i))^2+\lambda(\bm{g}\odot\mathbf{x}_i)^{\top}\mathbf{x}_i\right\} = \bm{0},\\
    \frac{1}{n} \sum_{1}^{n}& -2(\bm{\beta}^{\top}(\bm{q}\odot\mathbf{x}_i))(\bm{\beta}\odot\mathbf{x}_i) + \lambda \mathbf{x}_i^2 = \bm{0},
\end{align*}
where $\mathbf{x}_i^2$ denotes a vector whose elements are $x_j^2$. Given that $\mathbf{x}$ is zero mean with covariance matrix $\bm{C}_n$, defining $\bm{q=1-g}$ and simple algebraic reordering results in 
\begin{equation*}
    \frac{1}{n} \sum_{1}^{n} (\bm{q}^{\top}(\bm{\beta}\odot\mathbf{x}_i))(\bm{\beta}\odot\mathbf{x}_i) = \left(\lambda/2\right)\mathrm{diag}(\bm{C}_n)
\end{equation*}
Reformulating the above equation as a linear system of equations in terms of $\bm{q}$, we obtain the following solution for the unconstrained problem:
\begin{align*}
    \bm{g} = \bm{1}-\left(\lambda/2\right)(\bm{BC}_n\bm{B})^{-1}\mathrm{diag}(\bm{C}_n).
\end{align*}
The final result is the projection of the above solution to the valid solutions interval of $[0, 1]$. The projection into the $[0, 1]$ interval equals the clamping operation.
\end{proof}

%% file: details.tex
\paragraph{Implementation of the gating transformation:}
We use PyTorch \citep{paszke2017automatic} for implementing our algorithm. The transformation function is implemented as follows:

\texttt{Conv1d(in\_channels=34*e, 34, kernel\_size=1, groups=34, stride=1, padding=0, dilation=1, bias=True, padding\_mode='zeros')}

where $e=4$ if we use the basis functions and $e=1$ otherwise.

We use the proximal algorithm to enforce the condition on $\bm{g}$ variables by clamping the coefficients back into $[0, 1]$ after every optimization step.

\paragraph{Implementation of the residual transformation function:}
The residual transformation function consists of a cascade of two residual blocks in the form of \texttt{y = x + conv2(relu(bnorm(conv1(x))))}, where \texttt{conv1 = Conv1d(50, 150, kernel\_size=5, padding=2, bias=False)} and \texttt{conv2 = Conv1d(150, 50, kernel\_size=5, padding=2)}.

\paragraph{Implementation of the Transformer-based model:}

We first use \texttt{nn.Conv1d(34, 50, kernel\_size=5)} to embed the time series into a 50 dimensional space. We add positional encoding transformed by a linear function to the embedded input. The result is processed by \texttt{TransformerEncoderLayer(d\_model=50, nhead=5, dim\_feedforward=50, dropout=0.1)} with layer normalization after each layer. Finally, we use a \texttt{nn.Conv1d(50, 1, kernel\_size=1)} followed by global average pooling to estimate the logit.

\paragraph{Implementation of the Fully-Convolutional architecture in Section \ref{sec:sanity}:} We use 8 layers of \texttt{nn.Conv1d} with hidden sizes $60, 100, 120, 100, 80, 60, 40, 20$ and dilation factors $1, 2, 2, 2, 2, 2, 2, 1$. After every convolutions, we use \texttt{nn.BatchNorm1d} and GeLU nonlinearity function. For prediction, we use \texttt{nn.Linear(80, 1)} on the flattened output of the convolutions processed by another batch normalization and GeLU nonlinearity function.

%% file: names.tex
\begin{table*}[t]
\centering
\caption{Names of the variables used in the MIMIC-III benchmark \citep{harutyunyan2017multitask}}
\label{tab:names}
\begin{tabular}{lll}
\toprule
\textbf{Variable Index} & \textbf{Description} & \textbf{Abbreviation in Figs \ref{fig:regular} \& \ref{fig:adversarial}}\\
\midrule
1 & Capillary refill rate & Cap Refill \\
2 & Diastolic blood pressure & Diastolic BP\\
3 & Fraction inspired oxygen & $\mathrm{FiO}_2$\\
4 & Glascow coma scale eye opening & GCS eye\\
5 & Glascow coma scale motor response & GCS motor \\
6 & Glascow coma scale total & GCS total\\
7 & Glascow coma scale verbal response & GCS verbal\\
8 & Glucose & Glucose\\
9 & Heart Rate & Heart Rate\\
10 & Height & Height\\
11 & Mean blood pressure & Mean BP \\
12 & Oxygen saturation & $\mathrm{SpO}_2$\\
13 & Respiratory rate & Resp. Rate\\
14 & Systolic blood pressure & Systolic BP\\
15 & Temperature & Temperature\\
16 & Weight & Weight\\
17 &pH & PH\\
 \bottomrule
\end{tabular}
\end{table*}